\documentclass{article}

\usepackage{microtype}
\usepackage{graphicx}
\usepackage{subcaption}
\usepackage{booktabs}
\usepackage{hyperref}

 \usepackage[preprint]{icml2026}

\usepackage{amsmath}
\usepackage{amssymb}
\usepackage{mathtools}
\usepackage{amsthm}

\usepackage[capitalize,noabbrev]{cleveref}

\theoremstyle{plain}

\theoremstyle{definition}

\theoremstyle{remark}

\usepackage[textsize=tiny]{todonotes}

\icmltitlerunning{Plausible but Wrong: A Case Study on Agentic Failures in Astrophysical Workflows}

\begin{document}

\twocolumn[
\icmltitle{Plausible but Wrong: A Case Study on Agentic Failures in Astrophysical Workflows}

\begin{icmlauthorlist}
\icmlauthor{Shivam Rawat}{bonn}
\icmlauthor{Lucie Flek}{bonn}
\end{icmlauthorlist}

\icmlaffiliation{bonn}{Bonn-Aachen International Center for Information
Technology, University of Bonn, Germany;
Lamarr Institute for Machine Learning and Artificial Intelligence, Germany}

\icmlcorrespondingauthor{Shivam Rawat}{s.rawat@uni-bonn.de}

\icmlkeywords{LLM agents, scientific workflows, astrophysics,
Bayesian inference, evaluation, reliability}

\vskip 0.3in
]

\printAffiliationsAndNotice{}

\begin{abstract}
Agentic AI systems are increasingly being integrated into scientific
workflows, yet their behavior under realistic conditions remains
insufficiently understood. We evaluate CMBAgent across two workflow
paradigms and eighteen astrophysical tasks. In the One-Shot setting,
access to domain-specific context yields an approximately ${\sim}6\times$
performance improvement ($0.85$ vs.\ $\approx\!0$ without context), with
the primary failure mode being silent incorrect computation---syntactically
valid code that produces plausible but inaccurate results. In the Deep
Research setting, the system frequently exhibits silent failures across
stress tests, producing physically inconsistent posteriors without
self-diagnosis. Overall, performance is strong on well-specified tasks
but degrades on problems designed to probe reasoning limits, often
without visible error signals. These findings highlight that the most
concerning failure mode in agentic scientific workflows is not overt
failure, but confident generation of incorrect results. We release our
evaluation framework to facilitate systematic reliability analysis of
scientific AI agents.
\end{abstract}

\section{Introduction}

Recent advances in large language models (LLMs) \citep{brown2020language,
wei2022chain, achiam2023gpt} and tool-augmented reasoning systems
\citep{yao2022react, schick2023toolformer, nakano2021webgpt} have enabled
the emergence of agentic AI systems capable of multi-step reasoning,
external tool use, and autonomous decision-making. These systems are
increasingly proposed as assistants for scientific workflows, where they
can perform data analysis, hypothesis generation, and pipeline
orchestration with limited human intervention.

Despite promising demonstrations, existing work largely evaluates such
systems through task-level success or synthetic benchmarks, offering
limited insight into their reliability in realistic scientific settings.
Scientific workflows differ fundamentally from standard AI benchmarks:
they involve structured pipelines, domain-specific constraints, noisy
data, and multi-stage reasoning where early errors can propagate
throughout the pipeline. As a result, aggregate success metrics alone are
insufficient to characterise system behaviour in these environments.

Importantly, our focus is not on proposing or critiquing agentic
architectures themselves, but on evaluating their scientific reliability
and failure modes when deployed in realistic workflows.

A central challenge is that agentic systems inherit known limitations of
LLMs, including hallucination, brittle reasoning, and misalignment with
task objectives. When embedded in multi-step workflows with external tool
interaction, these issues can lead not only to incorrect intermediate
steps but also to silent numerical errors and physically inconsistent
outputs. Critically, such failures may not be detectable from final
outputs alone, making systematic evaluation of reliability and error
modes essential for scientific deployment.

In this work, we present a structured empirical evaluation of an existing
agentic system applied to astrophysical inference tasks. Rather than
introducing a new model or framework, our goal is to characterise
performance boundaries, assess robustness and calibration, and identify
systematic failure modes in realistic scientific workflows.

Astrophysics provides a particularly suitable testbed due to its
well-defined physical models, established numerical tools, and availability
of reference solutions from both simulation and observation. This allows
controlled evaluation of correctness while maintaining real-world
scientific relevance.

We introduce a structured evaluation framework for agentic scientific
workflows that integrates execution success, parameter accuracy, and
numerical fidelity. Using this framework, we conduct a systematic analysis
of reliability across both single-step and multi-step reasoning settings,
identifying key failure modes such as silent numerical errors and physically
inconsistent inference, and quantifying their prevalence under different
task conditions.

\section{Related Work}

\paragraph{LLM Agents and Tool Use.}
The ability of LLMs to interact with external tools has been a key
driver in the transition from static text generation to dynamic
agentic behaviour. ReAct \citep{yao2022react} established the
paradigm of interleaving reasoning traces with action execution,
enabling models to plan and adapt based on environmental feedback.
Toolformer \citep{schick2023toolformer} extended this by training
models to invoke APIs autonomously through self-supervision, while
WebGPT \citep{nakano2021webgpt} demonstrated that grounding language
models in live retrieval substantially improves factual reliability.
More recent work has formalised tool-use evaluation, with benchmarks
such as ToolBench \citep{qin2025meta} and TRAJECT-Bench
\citep{he2025traject} revealing systematic failure modes including
similar-tool confusion, parameter-blind selection, and performance
degradation with trajectory length. Together these works establish
that structured tool access, rather than parametric knowledge alone,
is what enables LLMs to operate effectively in multi-step
workflows --- a finding our results reinforce in the context of
scientific computation.

\paragraph{LLM Agents in Scientific Workflows.}
A growing number of works demonstrate that agentic systems can
execute substantial portions of real scientific workflows, spanning
hypothesis generation, data analysis, and manuscript production
\citep{ding2025scitoolagent, sun2025scienceboard,
villaescusa2025denario, moreno2026ai, hellert2025agentic}. While
these works establish the feasibility of agentic scientific
automation, evaluation is typically limited to end-to-end task
success or qualitative expert assessment. Systematic analysis of
reliability, failure modes, and error propagation in multi-step
scientific pipelines remains largely unexplored.

\paragraph{Reliability and Evaluation of LLMs.}
A substantial body of work examines the limitations of LLMs,
including hallucination \citep{huang2025survey, alansari2025hallucination},
factual inconsistency \citep{lin2022truthfulqa}, and misalignment
with user intent \citep{bai2022training}. Critically, recent surveys
show that hallucinations in agentic settings differ qualitatively
from single-step LLM failures: they emerge through multi-step
sequential interactions involving tool use, memory, and inter-agent
communication, making them substantially harder to detect from
final outputs alone \citep{lin2025agenthallu}. More recent work
has begun to formalise agent evaluation, distinguishing between
tool-use capability, trajectory-level reasoning, and domain-specific
performance \citep{yehudai2025survey, chowa2026language}. However,
systematic evaluation of reliability under realistic multi-step
scientific workflows remains limited.

\paragraph{Positioning of This Work.}
In contrast to prior work, we focus explicitly on reliability and
failure characterisation rather than task feasibility. Rather than
proposing a new model or architecture, we provide a structured
empirical evaluation of an existing deployed system under controlled
conditions such as under-constrained inference and compositional
reasoning pipelines. This enables us to identify not only whether
agentic systems succeed, but how and why they fail, and crucially,
whether those failures are visible in the final output or silent.

\section{Agentic System Under Consideration}

For our evaluation, we utilize \textit{CMBAgent}, an open-source agentic
framework for autonomous scientific discovery \citep{xu2025open}.
\textit{CMBAgent} serves as the research backend in the \textit{Denario}
project \citep{villaescusa2025denario}, generating research ideas,
methodologies, and results through coordinated multi-agent planning. The
outputs---including analyses, visualizations, and literature
references---are automatically compiled into publication-ready manuscripts.
While the \textit{Denario} project demonstrates the feasibility of
end-to-end autonomous scientific workflows and evaluates overall task
success through expert assessment, a systematic investigation of
reliability, robustness, error propagation, and failure modes remains
limited. Such analysis is essential for assessing the suitability of
agentic systems in scientific deployment settings where reproducibility and
numerical correctness are critical. Here we complement existing feasibility
studies by conducting a structured empirical evaluation focused on
robustness, calibration, and failure characterization, thereby contributing
toward a more rigorous evaluation framework for autonomous scientific
agents.

The framework provides multiple operational modes and workflows, which
differ primarily in the number of active agents, session length, context
persistence across tasks, and the extent of human intervention required
for conducting scientific analyses. These are as follows:

\begin{enumerate}
\item \textbf{One Shot.} Executes a task in a single reasoning and
execution pass without iterative planning or review. We instantiate
two variants: one with a context-aware domain agent
(\textit{cambagent}) providing task-relevant grounding, and one
without, isolating the contribution of domain-specific retrieval
augmentation relative to a base LLM baseline.

\item \textbf{Deep Research.} Activates the full Planning \&
Control architecture, decomposing tasks into substeps executed by
specialised agents with critique and retry modules. Designed for
complex, multi-step analyses requiring multilayered reasoning and
iterative error correction.
\end{enumerate}

We focus exclusively on these two fully automated workflows; the
\textbf{Human in the Loop} mode is excluded as user feedback introduces
variability that complicates reproducible benchmarking
\citep{yehudai2025survey, chowa2026language}.

\section{Evaluation and Metrics}

We evaluate the agentic system across two distinct categories of tasks,
aligned with the two workflow paradigms under investigation. This
separation is intentional and reflects the differing operational objectives
of the workflows. Tool-driven computational tasks are employed to evaluate
the \textit{One-Shot} workflow, as they emphasize single-pass reasoning,
precise tool invocation, and accurate parameter configuration. These tasks
allow controlled, fine-grained assessment of execution correctness and
numerical reliability without introducing the additional complexity of
multi-stage planning.

In contrast, the Deep Research workflow is specifically designed to support
iterative reasoning, hierarchical task decomposition, and sustained context
retention across multiple steps. Evaluating this workflow using simple,
single-call computational tasks would not meaningfully exercise its
planning and control mechanisms. Therefore, we assign more complex,
research-oriented tasks to the Deep Research workflow---tasks that require
multi-step reasoning, contextual carryover, and strategy
refinement---so that its architectural capabilities can be appropriately
tested. This task--workflow alignment ensures that each paradigm is
evaluated under conditions that reflect its intended operational design.

Our evaluation design follows recent taxonomies of LLM-agents
\citep{yehudai2025survey}, which distinguish between capability-centric
evaluation (e.g., tool use and function correctness), trajectory-level
evaluation (e.g., planning and multi-step reasoning), and domain-specific
application evaluation. The CAMB-based tasks primarily assess tool
invocation accuracy and computational reliability, whereas the
research-driven tasks stress hierarchical planning, long-horizon reasoning,
and statistical inference consistency in scientific workflows.

\subsection{Task Descriptions}

\paragraph{Tool-Grounded Precision Tasks.}
Fourteen structured CAMB computation tasks assess parameter configuration
robustness, solver reliability, and numerical accuracy, adapted from the
CMBAgent benchmark repository \citep{cmbagent_benchmarks}. We introduce
a complexity stratification: Tasks 1--6 use a single API call; Task 7
adds tensor handling; Task 8 targets an alternate CAMB module; Tasks 9--10
require multi-API calls with combined results; Tasks 11--12 add a
delensing pipeline; Task 13 adds ratio computation; and Task 14 combines
multi-API calls with a noise-informed delensing pipeline.

\paragraph{Astrophysical Research-Driven Tasks.}
Four research-grade inference problems spanning cosmology, galactic
dynamics, exoplanet structure, and strong gravitational lensing probe
multi-step reasoning, statistical robustness, and resistance to
hallucinated physical assumptions via Bayesian parameter estimation and
hierarchical modeling. Task descriptions and category rationales are given
in Section~\ref{sec:ablations}; prompts are in the Appendix.

\subsection{Evaluation Metrics}
\label{sec:metrics}

We evaluate each workflow under a distinct scoring framework reflecting
its operational objectives. The One-Shot workflow admits fully automated
quantitative evaluation against reference outputs; the Deep Research
workflow requires qualitative assessment against published literature
values. All metrics are scored in $[0,1]$ and averaged across trials
for reporting.

\subsubsection{One-Shot Workflow Metrics}

\paragraph{Execution Success Rate (ESR).}
ESR is a binary indicator of whether the agent produced valid executable
output. A trial is assigned $\mathrm{ESR} = 1$ if and only if the agent
generates a numerical output file that (i)~contains at least two numeric
columns, (ii)~covers at least 95\% of the reference $x$-range, and
(iii)~provides at least 95\% of the reference number of output points.
Otherwise $\mathrm{ESR} = 0$. This criterion deliberately does not assess
numerical correctness---it only verifies that the pipeline completed and
produced an output of the right shape.

\paragraph{Parameter Accuracy Score (PAS).}
PAS measures whether the agent configured the CAMB cosmological solver
with the correct input parameters. Parameters are extracted directly from
the generated code via abstract syntax tree (AST) parsing, without
executing the code, making the metric robust to runtime failures. For each
parameter $p$ present in the reference solution, the relative error is
\begin{equation}
    \epsilon_p = \min\!\left(
        \frac{|\hat{\theta}_p - \theta_p^*|}{|\theta_p^*| + \delta},\; 1
    \right),
\end{equation}
where $\theta_p^*$ is the reference value, $\hat{\theta}_p$ is the
agent's value, and $\delta = 10^{-12}$ prevents division by zero. If the
agent omits a required parameter, $\epsilon_p = 1$. The overall PAS is a
weighted mean over all reference parameters:
\begin{equation}
    \mathrm{PAS} = 1 - \frac{\sum_p w_p\, \epsilon_p}{\sum_p w_p},
\end{equation}
where weights $w_p$ reflect the physical importance of each parameter:
$w = 2.0$ for $\{H_0, \Omega_b h^2, \Omega_c h^2, n_s, A_s\}$,
$w = 1.5$ for $\{\Omega_k, w_0\}$, and
$w = 1.0$ for $\{\tau, m_\nu, A_\mathrm{lens}\}$. Weights reflect the
breadth of spectral impact: core $\Lambda$CDM parameters ($H_0$,
$\omega_b h^2$, $\omega_c h^2$, $n_s$, $A_s$; $w=2.0$) distort
amplitude, shape, and peak positions globally; extension parameters
($\Omega_k$, $w_0$; $w=1.5$) are task-critical only when non-zero; and
secondary parameters ($\tau$, $m_\nu$, $A_\mathrm{lens}$; $w=1.0$) have
localised effects. This scheme underpenalises extension-parameter errors
on tasks where they are non-trivial, making reported PAS a slight
overestimate in those cases.

\paragraph{Numerical Accuracy Score (NAS).}
NAS measures how closely the agent's output curve matches the reference
solution over the full output domain. The agent's curve is interpolated
onto the reference grid and three complementary sub-metrics are computed.

\textit{NRMSE score} measures amplitude accuracy:
\begin{equation}
    S_\mathrm{NRMSE} = \max\!\left(0,\; 1 -
        \frac{\sqrt{\frac{1}{N}\sum_i (\hat{y}_i - y_i^*)^2}}
             {\max(y^*) - \min(y^*)}
    \right).
\end{equation}

\textit{SMAPE score} measures symmetric percentage deviation,
robust to scale differences:
\begin{equation}
    S_\mathrm{SMAPE} = \max\!\left(0,\; 1 -
        \frac{1}{N}\sum_i
        \frac{|\hat{y}_i - y_i^*|}
             {(|\hat{y}_i| + |y_i^*|)/2}
    \right).
\end{equation}

\textit{CCC score} is Lin's concordance correlation coefficient
\citep{lin1989concordance}, which jointly penalises location shift,
scale shift, and correlation:
\begin{equation}
    S_\mathrm{CCC} = \max\!\left(0,\;
        \frac{2\,\mathrm{Cov}(\hat{y}, y^*)}
             {\mathrm{Var}(\hat{y}) + \mathrm{Var}(y^*) +
              (\bar{\hat{y}} - \bar{y}^*)^2}
    \right).
\end{equation}

The sub-scores are combined as:
\begin{equation}
    \mathrm{NAS} = 0.2\, S_\mathrm{NRMSE}
                 + 0.3\, S_\mathrm{SMAPE}
                 + 0.5\, S_\mathrm{CCC}.
\end{equation}
\textit{CCC} receives the highest weight because it simultaneously captures
amplitude, scale, and shape agreement; \textit{NRMSE} is down-weighted
because it is sensitive to outliers and can be dominated by a single
high-amplitude peak.

\paragraph{Final Score.}
\begin{equation}
    \mathrm{Score} =
    \begin{cases}
        0 & \text{if } \mathrm{ESR} = 0, \\
        \mathrm{PAS} \times \mathrm{NAS} & \text{otherwise.}
    \end{cases}
\end{equation}
The multiplicative form ensures that a trial scores zero if either
parameters or numerical output are entirely wrong, even if the other
dimension is perfect.

\paragraph{Failure Mode Taxonomy.}
Each trial is assigned to one of four mutually exclusive failure modes:
\begin{itemize}
    \item[\textbf{A}] \textbf{Code failure} ($\mathrm{ESR} = 0$):
          the pipeline did not produce valid output.
    \item[\textbf{B}] \textbf{Wrong parameters}
          ($\mathrm{ESR} = 1$, $\mathrm{PAS} < 0.5$):
          the code ran but used incorrect CAMB configuration.
    \item[\textbf{C}] \textbf{Wrong computation}
          ($\mathrm{ESR} = 1$, $\mathrm{PAS} \geq 0.5$,
           $\mathrm{NAS} < 0.5$):
          parameters were correct but numerical output was
          inaccurate, indicating a formula, unit, or
          post-processing error.
    \item[\textbf{D}] \textbf{Correct}
          ($\mathrm{ESR} = 1$, $\mathrm{PAS} \geq 0.5$,
           $\mathrm{NAS} \geq 0.5$):
          both configuration and numerical output are
          consistent with the reference.
\end{itemize}
We additionally flag \textit{unit/normalisation errors} as a sub-class
of Mode~C: trials where $S_\mathrm{CCC} > 0.8$ but
$S_\mathrm{NRMSE} < 0.7$ indicate correct spectral shape with wrong
amplitude, consistent with a missing normalisation factor or unit
conversion error.

\subsubsection{Deep Research Workflow Metrics}

The Deep Research workflow produces no automated reference CSV, so
ESR, PAS, and NAS are not applicable. Each task is instead evaluated
across two dimensions: a quantitative parameter recovery score and a
qualitative physical consistency assessment.

\paragraph{Parameter Recovery Score (PRS).}
Measures agreement between the agent's reported posterior median and
the literature reference for each key parameter. For parameter $p$
with reference $\theta_p^* \pm \sigma_p^*$, the per-parameter score
is:
\begin{equation}
    r_p = \max\!\left(0,\; 1 - \frac{|\hat{\theta}_p -
    \theta_p^*|}{3\,\sigma_p^*}\right),
\end{equation}
saturating to zero at $3\sigma$ deviation. PRS is the unweighted
mean of $r_p$ over all task parameters.

\paragraph{Qualitative Assessment.}
Beyond parameter recovery, each task is assessed along two
dimensions reported as free-text observations in the results:
(i)~\textit{physical plausibility} --- whether reported parameter
values and posterior shapes are consistent with established physical
expectations for the system; (ii)~\textit{failure transparency} ---
whether the agent identifies and reports known degeneracies,
structural biases, or convergence issues, rather than delivering
results silently --- reported alongside PRS in the results.

\subsection{Task-Level Robustness Evaluation}
\label{sec:ablations}

To evaluate reliability beyond aggregate performance, we design the
research-driven task suite as a structured stress test of the Deep
Research workflow. Rather than ablating architectural components, we
ablate the \textit{task conditions} themselves---varying the degree of
prior specification, likelihood completeness, and model
complexity---while keeping the workflow fixed. This approach is
motivated by prior observations that agentic failures in scientific
settings emerge not from random errors but from systematic weaknesses
under specific inference regimes: two structurally distinct
under-constrained posteriors and compositional reasoning chains
\citep{yehudai2025survey, chowa2026language}.

We evaluate the One-Shot workflow through a controlled architectural
ablation: access to domain-specific CAMB documentation is removed in
the \textit{CMBAgent (no context)} variant, isolating the contribution
of structured retrieval augmentation on tool invocation accuracy and
parameter configuration relative to both the full CMBAgent system and
the direct Base LLM baseline.

The four research-driven tasks are assigned to these conditions as
follows:

\begin{itemize}

\item \textbf{Under-Constrained Inference Stress Tests --- Tasks 1, 2}

  \textbf{Task 1:} Fitting the Union2.1 Type Ia supernova
  distance--redshift data with a flat $\Lambda$CDM cosmology using
  MCMC to jointly estimate $H_0$ and $\Omega_\Lambda$. The SN1a
  likelihood is degenerate in $H_0$ and the supernova absolute
  magnitude $M_B$; without an independent distance anchor, $H_0$
  cannot be constrained by this dataset alone.

  \textbf{Task 2:} Modeling the rotation curve of NGC\,3198 by
  combining stellar, gaseous, and bulge contributions with an NFW
  dark matter halo profile to infer halo virial mass $M_{200}$ and
  concentration $c$, including uncertainty quantification via MCMC.

  \textbf{Category Rationale:}
  Both tasks probe silent over-confidence under known parameter
  degeneracies: the $H_0$--$M_B$ degeneracy in T1 and the
  mass--concentration degeneracy in T2. A reliable agent should flag
  these as prior-dominated rather than report confident but
  physically uninformative constraints.

  \textbf{Task label (T1):} \textit{Under-constrained} --- flat
  prior on $H_0$, $H_0$--$M_B$ degeneracy unbroken by design.

  \textbf{Task label (T2):} \textit{Under-constrained} --- flat
  priors on $(\log_{10} M_{200},\, c)$, mass--concentration
  degeneracy unbroken by design.

\item \textbf{Compositional Bayesian Workflow Evaluation --- Tasks 3, 4}

  \textbf{Task 3:} Performing a hierarchical MCMC analysis of the
  exoplanet mass--radius relation for planets with $M > 2\,M_{E}$,
  incorporating intrinsic scatter, measurement uncertainties, and
  temperature-dependent radius inflation, with comparative analysis
  of Neptunian and Jovian regimes.

  \textbf{Task 4:} Conducting Bayesian inference for SLACS
  strong-lensing systems using a Singular Isothermal Sphere (SIS)
  model to recover posterior velocity dispersions and compare
  lensing-inferred values to observed stellar velocity dispersions.

  \textbf{Category Rationale:}
  Both tasks test silent inconsistency rather than degeneracy
  detection: whether the agent sustains physical consistency and
  stable parameterisation across a multi-stage pipeline with no
  artificial degradation. Failures here manifest as inter-trial
  drift, prior boundary pathologies, or population-level bias ---
  none of which trigger obvious error signals.

  \textbf{Task labels (T3, T4):} \textit{Compositional} --- full
  hierarchical likelihood, no prior degradation.

\end{itemize}

\section{Experiments}
\label{sec:experiments}

\subsection{Systems}
\label{sec:systems}

We evaluate three systems under the One-Shot workflow. The two
CMBAgent variants are evaluated under their default configuration;
full details of the agent-to-model assignments are provided in the
CMBAgent repository \citep{xu2025open}. The Base LLM baseline
uses a direct single-turn call to \texttt{GPT-4o-mini} with no
surrounding architecture, providing a clean lower bound on
performance attributable to the model alone.

\begin{itemize}
    \item \textbf{Base LLM.} A direct single-turn call to
    \texttt{GPT-4o-mini} with no agentic framework, no tool
    integration, and no CAMB documentation. The model receives
    the task prompt and must produce executable Python code in
    one pass.

    \item \textbf{CMBAgent (no context).} The CMBAgent One-Shot
    workflow under default configuration, with the
    \texttt{engineer} agent and up to 50 reasoning rounds,
    but without access to CAMB documentation.

    \item \textbf{CMBAgent (CAMB context).} The CMBAgent One-Shot
    workflow under default configuration, with the
    \texttt{camb\_context} agent providing retrieval access to
    the CAMB API documentation during execution. This is the
    full intended deployment mode of the system.
\end{itemize}

The Deep Research workflow is additionally applied to the four
Astrophysical Research-Driven Tasks (T1--T4); its configuration
and evaluation protocol are described in Section~\ref{sec:protocol}.

\subsection{Protocol}
\label{sec:protocol}

\paragraph{Trials.}
The One-Shot workflow is evaluated over $N = 10$ independent trials
per task--system combination, yielding $14 \times 10 \times 3 = 420$
total trials. All Base LLM calls use sampling temperature $T = 0.2$.
The Deep Research workflow is evaluated with $N=5$ independent trials
per task; evaluation of this workflow is therefore qualitative, based
on parameter recovery and physical consistency of the reported outputs
rather than statistical aggregation across trials.

\paragraph{Ground truth.}
For the One-Shot workflow, reference outputs are generated once per
task by executing the reference implementation from the CMBAgent
benchmark repository \citep{cmbagent_benchmarks} and saving the
resulting $(x, y)$ numerical array to a standardised CSV file.
For the Deep Research workflow, no automated reference CSV exists;
ground truth is instead taken from the published literature values
listed in Table~\ref{tab:dr_references}, covering all four
research-driven tasks (T1--T4).

\begin{table}[h]
\centering
\caption{Literature ground-truth values for Deep Research evaluation.}
\label{tab:dr_references}
\scriptsize
\setlength{\tabcolsep}{4pt}
\begin{tabular}{llll}
\toprule
\textbf{Task} & \textbf{Param.} & \textbf{Reference} & \textbf{Source} \\
\midrule
T1 & $\Omega_\Lambda$ & $0.72{\pm}0.02$ & \citep{suzuki2012hubble} \\
\midrule
T2 & $\log_{10}\!\left(\frac{M_{200}}{M_\odot}\right)$ & $11.97{\pm}0.15$ & \citep{karukes2015dark} \\
   & $c$                & $6.8{\pm}2.0$    & \citep{karukes2015dark} \\
\midrule
T3 & $\alpha_N$       & $0.67{\pm}0.05$  & \citep{muller2024mass} \\
   & $\alpha_J$       & $-0.06{\pm}0.07$ & \citep{muller2024mass} \\
   & $M_\mathrm{br} \; [M_\oplus]$  & $127{\pm}17$ & \citep{muller2024mass} \\
\midrule
T4 & $f$              & $1.019{\pm}0.008$ & \citep{bolton2008sloan} \\
\bottomrule
\end{tabular}
\end{table}

\paragraph{Output standardisation.}
To ensure comparable output formats across systems, the target output
file path is appended to every prompt at runtime. This eliminates
ambiguity in output location without modifying the scientific content
of the task prompt.

\paragraph{Deep Research configuration.}
Each Deep Research trial is allocated up to 50 planning rounds,
100 control rounds, and a maximum of 5 execution retries per
sub-step (\texttt{max\_n\_attempts=5}). The retry budget allows the
workflow to self-recover from environment compatibility errors such
as deprecated API calls, which were the dominant failure mode
observed across tasks.

\paragraph{Evaluation.}
For the One-Shot workflow, each trial output is evaluated against
the reference CSV using the ESR, PAS, NAS, and failure mode metrics
defined in Section~\ref{sec:metrics}. Per-task scores are averaged
across 10 trials; per-system scores are averaged across tasks and
trials. For the Deep Research workflow, evaluation is conducted by
comparing the agent's reported posterior medians against the
literature values in Table~\ref{tab:dr_references}, with success
additionally confirmed by the presence of output files beyond the
execution log.

\section{Results}
\label{sec:results}

\subsection{One-Shot Workflow}

\begin{figure}[h]
  \centering
  \includegraphics[width=\columnwidth]{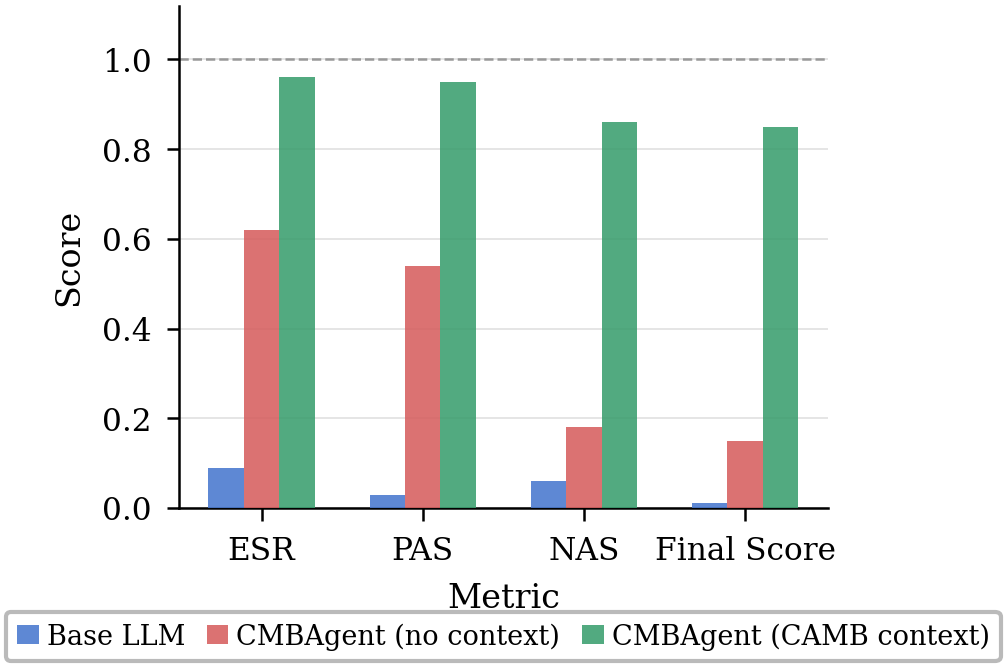}
  \caption{Mean ESR, PAS, NAS, and Final Score per system
           across 14 tasks and 10 trials.}
  \label{fig:system_scores}
\end{figure}

Figure~\ref{fig:system_scores} summarises mean ESR, PAS, NAS, and
Final Score across all 14 tasks and 10 trials per system.

\paragraph{CMBAgent with CAMB context.}
The full system achieves near-ceiling performance: ESR\,$=0.96$,
PAS\,$=0.95$, NAS\,$=0.86$, Final Score\,$=0.85$. The small gap
between PAS and NAS reflects residual numerical errors on the
hardest tasks even when parameters are correctly configured.

\begin{figure}[h]
  \centering
  \includegraphics[width=\columnwidth]{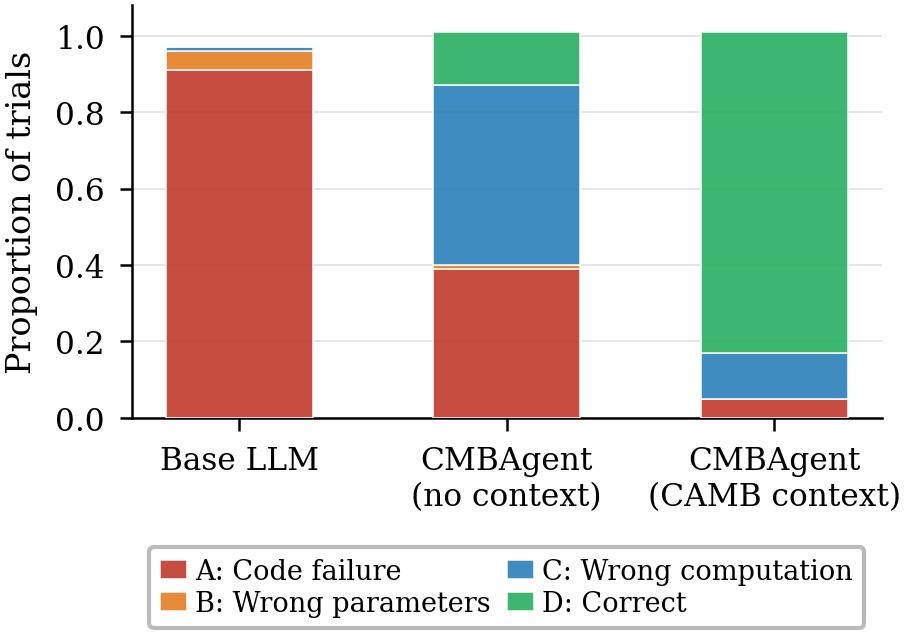}
  \caption{Failure mode breakdown per system as proportion of trials.}
  \label{fig:failure_modes}
\end{figure}

\paragraph{CMBAgent without context.}
Removing documentation access produces sharp degradation across all
metrics (ESR\,$=0.62$, PAS\,$=0.54$, NAS\,$=0.18$, Final\,$=0.15$).
The most severe drop is in NAS, consistent with
Figure~\ref{fig:failure_modes} where Mode~C (wrong computation)
accounts for $\approx\!47\%$ of trials --- the agent invokes
plausible but incorrect API calls, producing curves with the right
shape but wrong amplitude or spectral content.

\paragraph{Base LLM.}
The direct \texttt{GPT-4o-mini} baseline fails on virtually all
trials (ESR\,$=0.09$, Final\,$\approx\!0$), with $\approx\!91\%$
classified as Mode~A (code failure). Raw model capability without
agentic scaffolding is insufficient for tool-grounded scientific
computation.

\begin{figure*}[t]
  \centering
  \includegraphics[width=\textwidth]{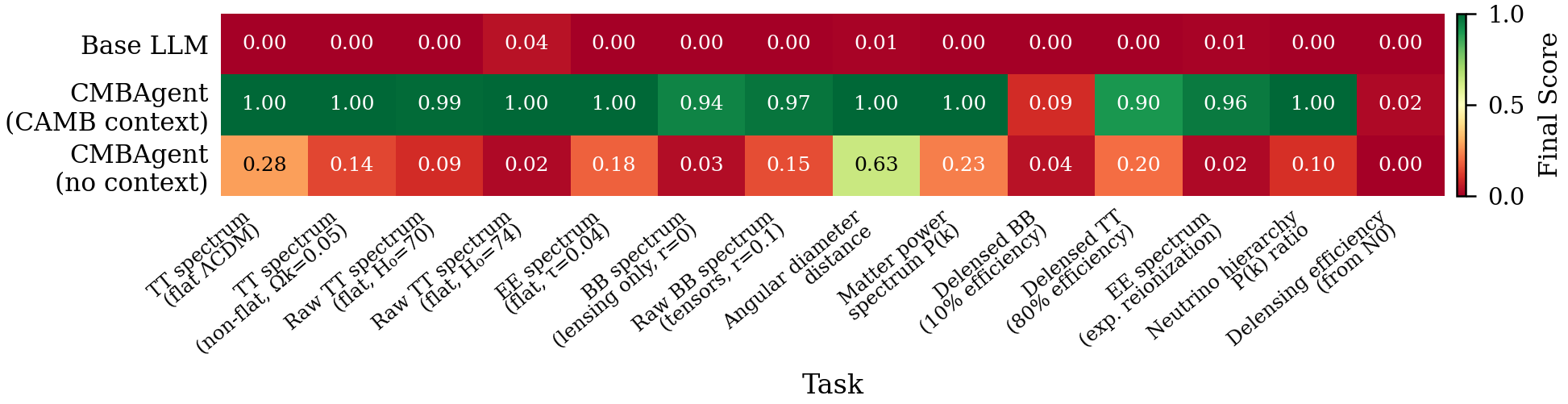}
  \caption{Final Score per system and task, averaged over 10 trials.
           Green indicates high score, red indicates low score.}
  \label{fig:heatmap}
\end{figure*}

\paragraph{Task-level analysis.}
Figure~\ref{fig:heatmap} breaks performance down across all 14 tasks.
CMBAgent with CAMB context scores near $1.0$ on most tasks but fails
silently on two: Task~10 (score $0.09$), where most trials omit
\texttt{raw\_cl=True} producing a ${\sim}10^5$ amplitude error and
all trials omit the tensor B-mode contribution; and Task~14 (score
$0.02$), where the agent correctly computes per-$\ell$ delensing
efficiency but collapses it to a scalar mean, yielding
\textit{CCC}\,$\approx 0$. Both failures stem from output
construction errors rather than incorrect API usage --- the
documentation does not prescribe unit conventions or post-processing
for these quantities. CMBAgent without context shows an isolated
peak at Task~8 ($0.63$), whose calling convention is common enough
to be known without documentation, but scores below $0.30$
elsewhere. The Base LLM scores near zero throughout.

\subsection{Deep Research Workflow}
\label{sec:dr_results}

The Deep Research workflow was evaluated on four research-grade tasks
(T1--T4) with up to five independent trials per task. Each task is
reported with its PRS score and a qualitative assessment across the
two dimensions defined in Section~\ref{sec:metrics}: physical
plausibility and failure transparency. Table~\ref{tab:dr_scores}
summarises results.

\paragraph{T1 --- SN1a Under-Constrained.}
All four completed trials recovered $\Omega_\Lambda$ within $0.15\sigma$
of the Union2.1 reference $0.720\pm0.020$ \citep{suzuki2012hubble},
with mean posterior $0.722\pm0.001$ confirming high reproducibility
($\mathrm{PRS}=0.97$). However, all trials deviated from the task
specification by treating $H_0$ as a free parameter alongside
$\Omega_\Lambda$. This mirrors a realistic failure scenario: the
prompt was intentionally constructed as a naive but plausible
scientific request, and a reliable agent should have identified the
$H_0$--$M_B$ degeneracy and flagged that $H_0$ cannot be constrained
from this dataset alone. Instead, the recovered $H_0$ posterior
simply reflects the flat prior $H_0\in[50,90]$ rather than any data
constraint, and was reported as a genuine measurement.

\textit{Physical plausibility:} $\Omega_\Lambda$ posteriors are
well-constrained and physically reasonable; the $H_0$ posterior is
prior-dominated and physically uninformative.

\textit{Failure transparency:} two silent failures on completed
trials --- the unauthorised addition of $H_0$ as a free parameter
yielding a prior-dominated posterior, and a runtime constraint
violation --- neither flagged by the agent.

\paragraph{T2 --- NGC\,3198 Under-Constrained.}
All five trials failed to recover parameters consistent with the
reference value ($\mathrm{PRS}=0.05$). Mass estimates spanned
$\log_{10}(M_{200}/M_\odot)\in[10.0,12.9]$ --- nearly three orders
of magnitude --- with no trial within $5\sigma$ of the reference
$11.97\pm0.15$ \citep{karukes2015dark}. Three trials produced
unphysical NFW concentrations: two high-mass trials
($\log_{10}M_{200}\approx12.8$--$12.9$) returned $c<2$, while one
trial hit the upper prior boundary at $c\approx30$.

\textit{Physical plausibility:} three of five trials returned
concentrations outside $c\in[2,30]$; the unphysical $c<2$ cases
coincided with high virial mass, tracing a degenerate
high-mass, low-concentration ridge.

\textit{Failure transparency:} one trial noted elongated contours
but no trial flagged unphysical posteriors or added caveats to
its parameter estimates.

\paragraph{T3 --- Exoplanet Mass--Radius Compositional.}
The Neptunian slope $\alpha_N\approx0.60$ is consistently recovered
across all four trials ($\sim1.4\sigma$ from reference
$0.67\pm0.05$, \citep{muller2024mass}). The Jovian slope $\alpha_J$
and break mass $M_\mathrm{br}$ are inconsistent across trials:
Trial~0 recovers $M_\mathrm{br}\approx127\,M_\oplus$ (within
$1\sigma$ of reference) with $\alpha_J=0.002$ ($\sim0.9\sigma$);
Trial~1 returns $\alpha_J=0.368$ ($6.1\sigma$ from reference
$-0.06\pm0.07$) and $M_\mathrm{br}\approx203\,M_\oplus$ ($4.5\sigma$
above reference); Trial~3 adopts a structurally different
parameterisation with $\log M_b=4.94$ on an inconsistent unit scale,
precluding direct comparison; Trial~4 recovers
$M_\mathrm{br}\approx118\,M_\oplus$ ($0.6\sigma$ below reference)
with $\alpha_J\approx0.000$ ($0.9\sigma$), but with
$\sigma_\mathrm{int}$ (the intrinsic scatter in the mass--radius
relation beyond measurement noise) pinned at the prior boundary
($0.5$), indicating inadequate prior specification. Mean PRS over
the three key parameters gives $\mathrm{PRS}=0.73$, driven by robust
$\alpha_N$ recovery but large variance in $\alpha_J$ and
$M_\mathrm{br}$.

\textit{Physical plausibility:} $\alpha_N$ is consistently
physical; $\alpha_J$ is wrong in sign or magnitude in three of
five trials; $M_\mathrm{br}$ is implausible in at least two trials
(Trial~1 at $4.5\sigma$ above reference; Trial~3 on an inconsistent
unit scale); $\sigma_\mathrm{int}$ hits the prior boundary in
Trial~4.

\textit{Failure transparency:} no trial flagged the inconsistent
model parameterizations, implausible Jovian slopes, discrepant
break masses, or the prior-boundary pathology in
$\sigma_\mathrm{int}$; confident estimates were reported throughout
without caveats.

\paragraph{T4 --- SLACS Strong Lensing Compositional.}
Only one of five trials produced usable output; the remaining four
failed to complete. The single completed trial yielded physically
coherent results. However, lensing-inferred dispersions are
systematically ${\approx}5\%$ above observed values, yielding
$f\approx1.05$ against the reference $f=1.019\pm0.008$
\citep{bolton2008sloan}, a ${\approx}4\sigma$ bias
($\mathrm{PRS}=0.00$).

\textit{Physical plausibility:} velocity dispersions and Einstein
radius predictions are within the physically expected range and the
SIS model is correctly implemented.

\textit{Failure transparency:} the systematic offset in $f$ was not
flagged by the agent; the four failed trials produced no output or
error diagnosis.

\begin{table}[h]
\centering
\caption{Deep Research results. PRS is computed from posterior
         medians vs.\ literature for completed trials only;
         qualitative dimensions
         (PP = physical plausibility, FT = failure transparency) are rated
         \checkmark\ (pass), $\sim$ (partial), or $\times$ (fail).}
\label{tab:dr_scores}
\scriptsize
\setlength{\tabcolsep}{3pt}
\begin{tabular}{llcccc}
\toprule
\textbf{Task} & \textbf{Category} & \textbf{Trials} & \textbf{PRS}
  & \textbf{PP} & \textbf{FT} \\
\midrule
T1 SN1a       & Under-constrained & 4/5 & 0.97 & $\sim$ & $\times$ \\
T2 NGC\,3198  & Under-constrained & 5/5 & 0.05 & $\times$ & $\times$ \\
T3 Exoplanets & Compositional     & 5/5 & 0.73 & $\sim$   & $\times$ \\
T4 SLACS      & Compositional     & 1/5 & 0.00 & \checkmark & $\times$ \\
\bottomrule
\end{tabular}
\end{table}

\section{Conclusion}

We presented a structured empirical evaluation of CMBAgent across two
workflow paradigms and eighteen tasks spanning tool-grounded computation
and research-grade Bayesian inference. Rather than focusing on aggregate
success, our goal was to characterise \textit{how} and \textit{why}
agentic systems fail, and whether those failures are detectable or silent.

Across both workflows, a consistent pattern emerges. In the One-Shot
setting, domain-context retrieval is the dominant performance driver:
with an identical \texttt{GPT-4o-mini} backbone, final score improves
from $\approx 0$ (Base LLM) to $0.15$ (no context) to $0.85$ (full
context), a ${\sim}6\times$ gain. Without context, the primary failure
mode is not execution failure but silent wrong computation---syntactically
valid code that produces plausible but incorrect results.

In the Deep Research setting, this failure pattern persists at the
inference level. Under-constrained tasks fail systematically:
degeneracies go undetected, yielding physically impossible posteriors,
yet results are reported as valid. Compositional workflows execute
successfully but exhibit inter-trial inconsistency, systematic bias,
and unstable parameterisations without caveats. Across all tasks,
outputs are consistently plausible, but failures remain unreported.
Failure transparency is the weakest dimension: the agent never
proactively flags known pathologies in its own outputs.

These findings highlight a critical risk for scientific deployment.
Agentic systems do not primarily fail by crashing---they fail by
producing confident, incorrect results or by silently breaking pipelines
without diagnosis. Our evaluation framework, combining quantitative
metrics for tool-grounded computation with structured qualitative
assessment for scientific inference, provides a template for
systematically identifying such failures. We argue that rigorous
reliability evaluation is a prerequisite for deploying agentic AI in
scientific workflows, where undetected errors can directly compromise
scientific conclusions.

\bibliography{example_paper}
\bibliographystyle{icml2026}

\appendix

\section{Deep Research Task Prompts}
\label{app:prompts}

\subsection{T1 --- SN1a Under-Constrained}
\label{app:prompt_t1}

\begin{quote}
Read the file:
\path{/home/sr/Desktop/code/cmbagent/cmbagent_systematics/deepresearch/task/SCPUnion2.1_mu_vs_z.txt}

Its description is: An ASCII table with tab-separated columns:
Supernova Name, Redshift, Distance Modulus, and Distance Modulus
Error. For Union2.1, there is an additional column for the
probability that the supernova was hosted by a low-mass galaxy.

Fit this data within a flat $\Lambda$CDM model with two free
parameters: $H_0$ and $\Omega_\Lambda$. Write a simple but
optimized MCMC to fit the SN1a data. Make a contour plot showing
the 1D posteriors and quote the mean and $1\sigma$ on each
parameter. Show the data alongside the best-fit model with 68\%
and 95\% CL regions. Comment on the results.

\textit{Constraints:} Running on a Dell Precision 5480 workstation
with 32\,GB RAM and 20 CPU threads (Intel Core i7-13800H). Use
resources optimally so the MCMC converges within a few minutes.
Have the engineer agent perform a preliminary MCMC timing step as
a separate preliminary step.
\end{quote}

\subsection{T2 --- NGC\,3198 Under-Constrained}
\label{app:prompt_t2}

\begin{quote}
Read the galaxy rotation curve file:
\path{/home/sr/Desktop/code/cmbagent/cmbagent_systematics/deepresearch/task/NGC3198_rotmod.txt}

\textit{Description:} SPARC (Spitzer Photometry and Accurate
Rotation Curves) provides high-quality rotation curves for nearby
disk galaxies. For NGC\,3198, the data includes: radius (kpc),
observed rotation velocity (km/s), velocity uncertainty, gas
contribution ($V_\mathrm{gas}$), disk contribution
($V_\mathrm{disk}$), and bulge contribution ($V_\mathrm{bul}$).
The observed rotation curve traces the total gravitational
potential.

Fit the rotation curve using an NFW (Navarro-Frenk-White) dark
matter halo model combined with the baryonic contributions. The
free parameters are: $M_{200}$ (virial mass of the dark matter
halo, in solar masses) and $c$ (concentration parameter).

Write an optimized MCMC code to fit these parameters. Show the 1D
posterior distributions and the 2D parameter contour plot. Quote
the mean and $1\sigma$ values.

Plot rotation velocity versus radius showing: observed data with
error bars, total best-fit model, individual components (disk, gas,
bulge, dark matter halo), and 68\% and 95\% confidence bands.

\textit{Constraints:} Run efficiently on a Dell Precision 5480
with 32\,GB RAM and 20 CPU threads. Ensure the MCMC completes
within a few minutes. Include a preliminary timing test.
\end{quote}

\subsection{T3 --- Exoplanet Mass--Radius}
\label{app:prompt_t3}

\begin{quote}
Read the exoplanet data file:
\path{/home/sr/Desktop/code/cmbagent/cmbagent_systematics/deepresearch/task/NASA_exoplanet_archive.csv}

\textit{Description:} The NASA Exoplanet Archive provides confirmed
exoplanet measurements. The file contains: planet name, mass ($M_p$
in $M_\oplus$) with uncertainties, radius ($R_p$ in $R_\oplus$)
with uncertainties, equilibrium temperature ($T_\mathrm{eq}$ in K)
with uncertainties, discovery method, stellar metallicity ([Fe/H])
with uncertainties, orbital period (days), semi-major axis (AU),
and system distance (pc). Filter to planets with $M_p > 2\,M_\oplus$
to focus on the Neptunian and Jovian regime.

Fit a broken power-law mass--radius relation with two regimes
separated by a break mass $M_\mathrm{br}$. Free parameters are:
Neptunian slope $\alpha_N$, Jovian slope $\alpha_J$, break mass
$M_\mathrm{br}$, normalisation, intrinsic scatter
$\sigma_\mathrm{int}$, and a temperature-dependent radius inflation
coefficient for hot planets.

Write an optimized MCMC sampler to obtain posterior distributions
of all parameters. Implement: data filtering above $2\,M_\oplus$;
hierarchical modelling of measurement uncertainties in both mass
and radius; intrinsic scatter perpendicular to the mass--radius
relation, and temperature-dependent corrections for radius inflation
in hot planets.

Produce the following plots: mass--radius diagram with best-fit
broken power-law curves; residuals versus mass showing the
transition at $M_\mathrm{br}$; posterior distributions for
$\alpha_N$, $\alpha_J$, and $M_\mathrm{br}$; a corner plot showing
parameter correlations; and a separate comparison of hot planets
($T_\mathrm{eq} > 1000$\,K) versus cold planets.

\textit{Constraints:} Dell Precision 5480, 32\,GB RAM, 20 CPU
threads. Target runtime 5--8 minutes. Use analytic power-law
models for efficient likelihood evaluation.
\end{quote}

\subsection{T4 --- SLACS Strong Lensing Compositional}
\label{app:prompt_t4}

\begin{quote}
Read the galaxy lensing data file:
\path{/home/sr/Desktop/code/cmbagent/cmbagent_systematics/deepresearch/task/Slac_data.csv}

\textit{Description:} The SLACS (Sloan Lens ACS) survey provides
high-quality data on strong gravitational lens systems. Columns
include: $z_d$ (lens redshift), $z_s$ (source redshift),
$R_\mathrm{eff}$ (effective radius, arcsec), $\theta_\mathrm{Ein}$
(Einstein radius, arcsec), $\sigma_\mathrm{obs}$ (stellar velocity
dispersion, km\,s$^{-1}$), and $\sigma_\mathrm{err}$
(km\,s$^{-1}$).

Fit the lens mass profile using a Singular Isothermal Sphere (SIS)
model with one free parameter per lens: the individual velocity
dispersion $\sigma_\mathrm{SIS}$. Use $z_d$ and $z_s$ to compute
angular diameter distances for each lens--source pair. Derive the
model-predicted Einstein radius from $\sigma_\mathrm{SIS}$ within
the SIS framework, and construct a likelihood comparing predicted
$\theta_\mathrm{Ein}$ to observed values for each lens
independently.

Write an optimized MCMC sampler to obtain the posterior
distribution of $\sigma_\mathrm{SIS}$ for each individual lens.

Produce the following plots: 1D marginalized posteriors for
representative lenses; a 2D comparison between lensing-inferred
$\sigma_\mathrm{SIS}$ and observed $\sigma_\mathrm{obs}$ (one
point per lens); model-predicted versus observed
$\theta_\mathrm{Ein}$; and the mass profile $M(<R)$ for each lens
overlaid with the Einstein radius scale.

\textit{Constraints:} Dell Precision 5480, 32\,GB RAM, 20 CPU
threads. MCMC should converge within ${\approx}5$ minutes. Use
vectorized likelihood evaluation across all lenses for efficiency.
\end{quote}

\section{Example Deep Research Outputs}
\label{app:results}

\subsection{T1 --- SN1a Under-Constrained (Representative Trial)}
\label{app:result_t1}

The completed trial recovers $\Omega_\Lambda = 0.722\pm0.019$,
consistent with the Union2.1 reference $0.720\pm0.020$
\citep{suzuki2012hubble}, but simultaneously reports
$H_0 = 69.99\pm0.34$\,km\,s$^{-1}$\,Mpc$^{-1}$ as a constrained
result. The $H_0$ posterior is prior-dominated: the Union2.1
likelihood cannot break the $H_0$--$M_B$ degeneracy without an
external distance anchor. The agent did not flag this, presenting
both parameters as equally reliable.

\begin{figure}[!htb]
  \centering
  \includegraphics[width=\columnwidth]{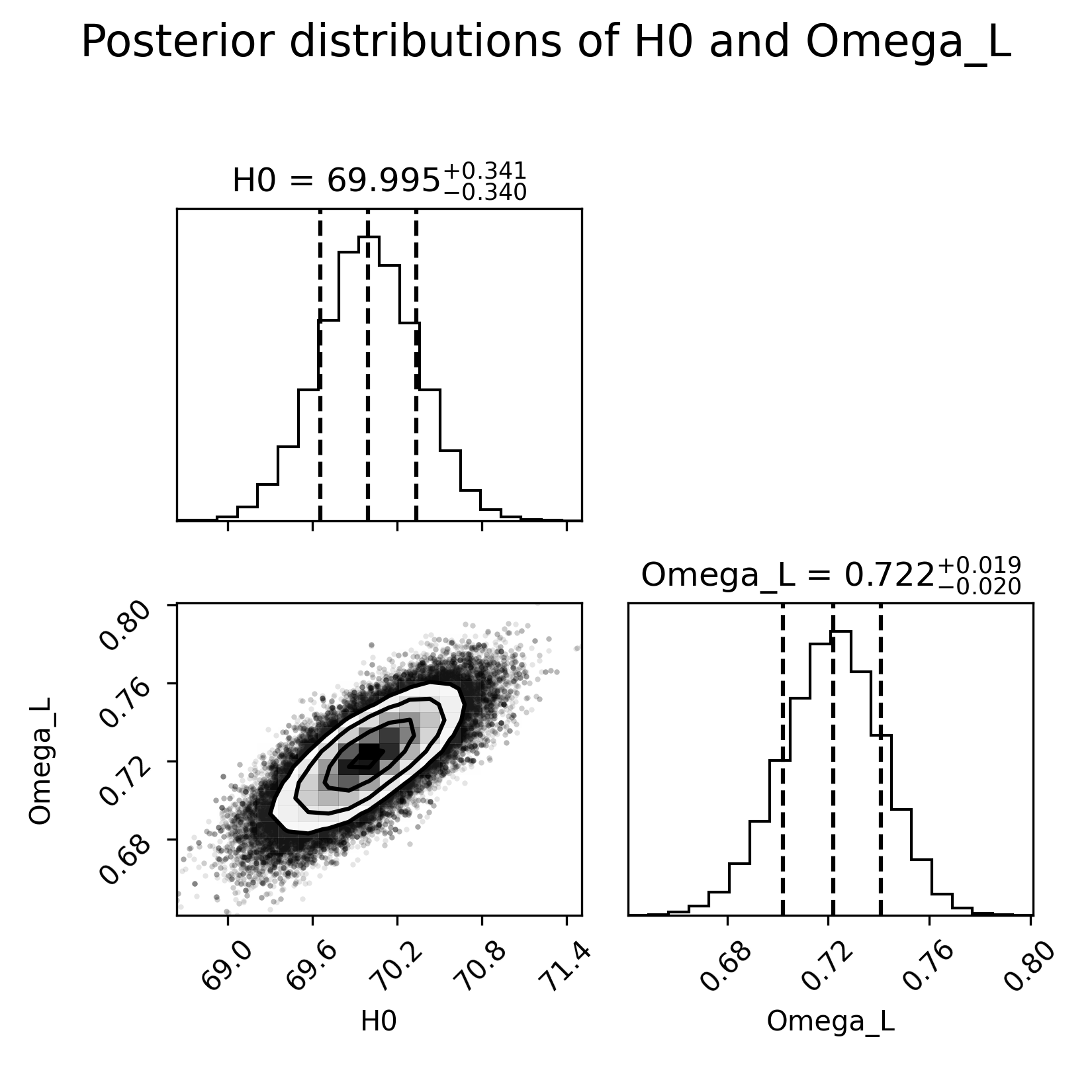}
  \caption{T1 posterior distributions of $H_0$ and $\Omega_\Lambda$
         from the Union2.1 SN1a fit. While $\Omega_\Lambda =
         0.722^{+0.019}_{-0.020}$ is well constrained and within
         $0.1\sigma$ of the reference, the $H_0$ posterior is
         prior-dominated --- the Union2.1 likelihood cannot
         break the $H_0$--$M_B$ degeneracy. The agent reported
         both parameters as equally reliable without flagging
         this pathology.}
  \label{fig:app_posterior_t1}
\end{figure}

\begin{figure}[!htb]
  \centering
  \includegraphics[width=\columnwidth]{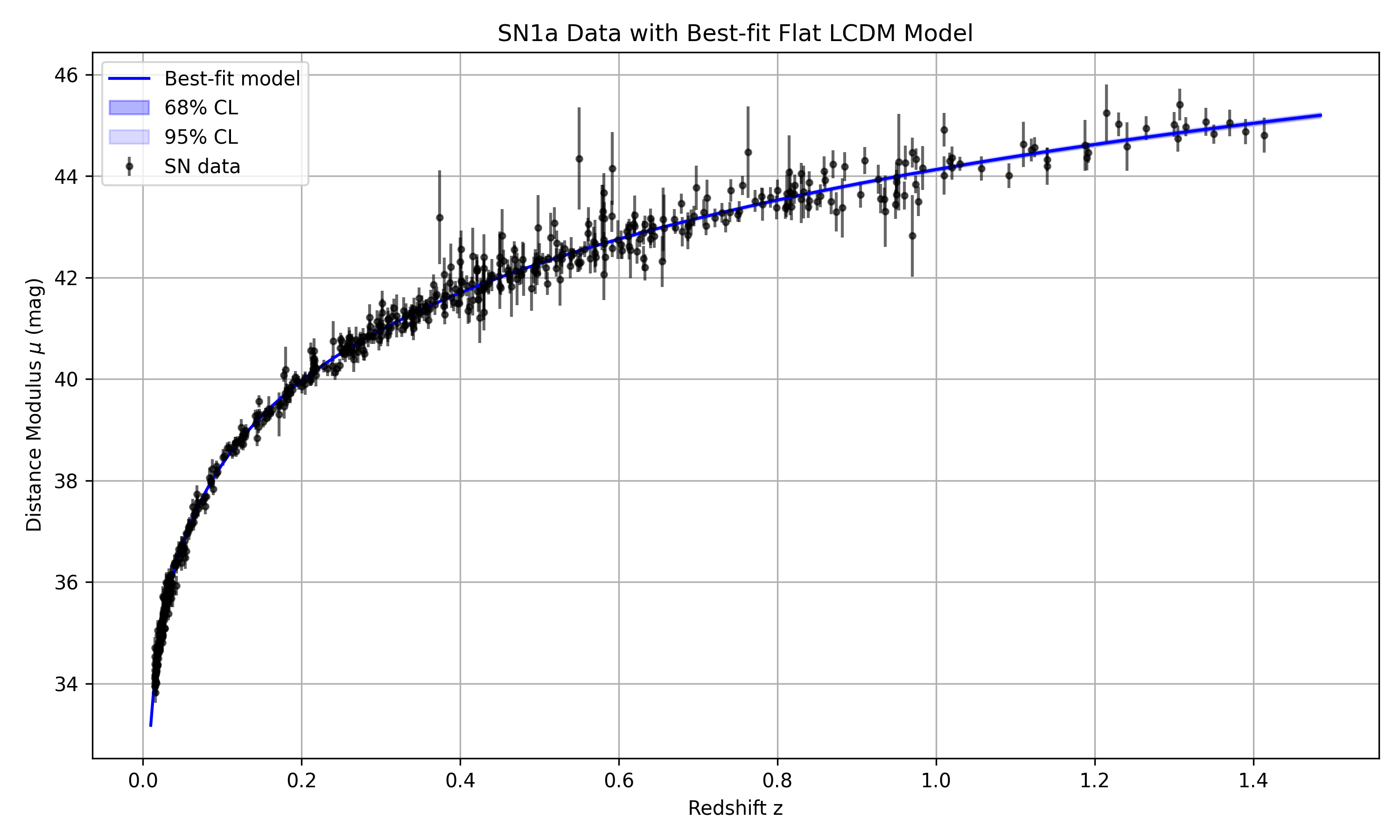}
  \caption{T1 best-fit flat $\Lambda$CDM model overlaid on
           Union2.1 SN1a data. The 68\% and 95\% CL bands
           are nearly indistinguishable from the best-fit
           curve, reflecting tight posterior constraints.}
  \label{fig:app_fit_t1}
\end{figure}

\subsection{T2 --- NGC\,3198 Under-Constrained (Failure Case)}
\label{app:result_t2}

This trial illustrates the primary failure mode for T2: the
posterior converges to $\log_{10}(M_{200}/M_\odot) =
12.90^{+0.06}_{-0.05}$, nearly $6\sigma$ above the reference
$11.97\pm0.15$ \citep{karukes2015dark}, with an unphysical
concentration $c = 1.19^{+0.21}_{-0.14} \ll 2$. The agent
reported these values without flagging the pathology.

\begin{figure}[!htb]
  \centering
  \includegraphics[width=\columnwidth]{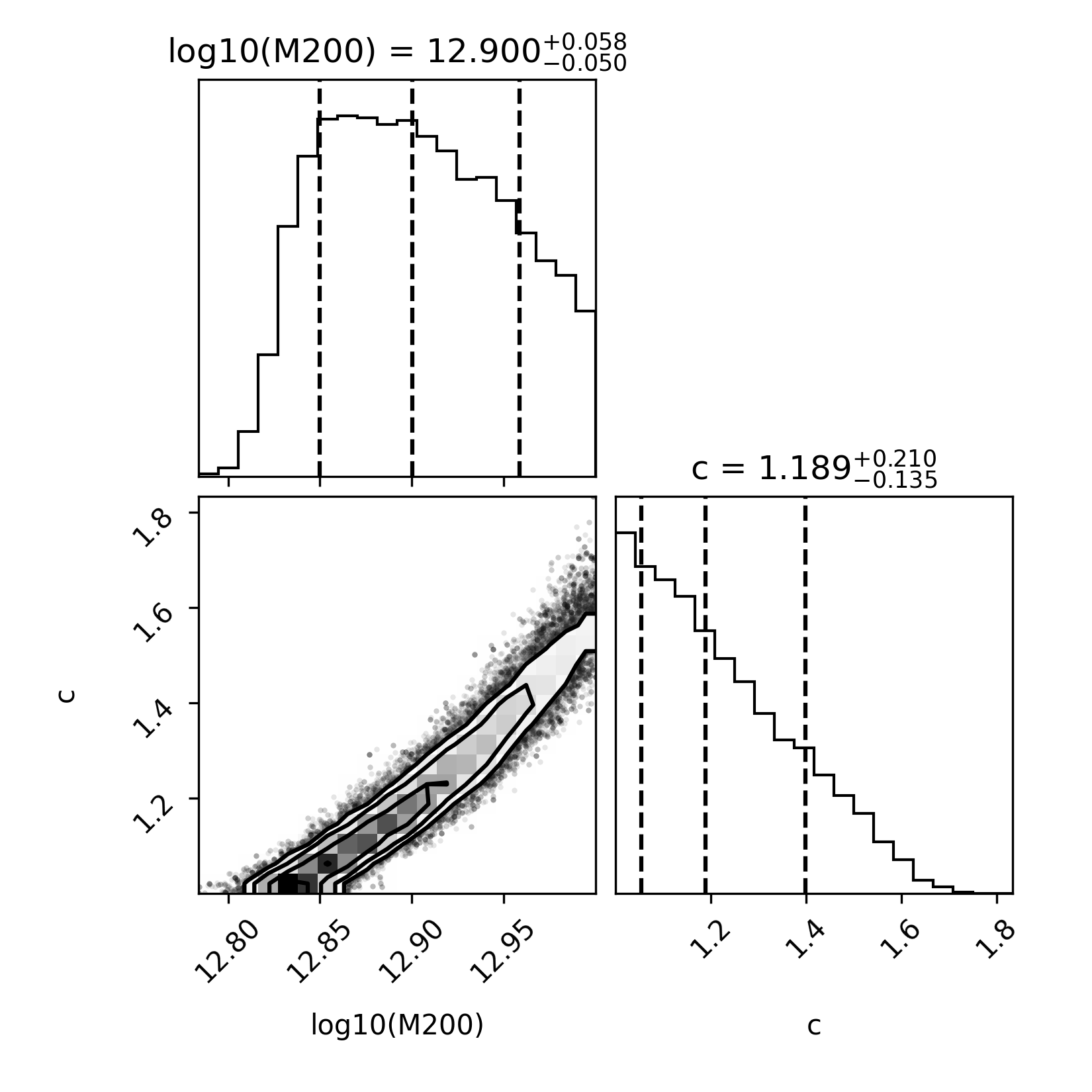}
  \caption{T2 posterior for $\log_{10}M_{200}$ and $c$,
           showing the degenerate high-mass, low-concentration
           ridge. The recovered $c \approx 1.2$ is physically
           impossible for an NFW halo; the reference value
           is $c = 6.8\pm2.0$ \citep{karukes2015dark}. No
           trial flagged this pathology.}
  \label{fig:app_posterior_t2}
\end{figure}

\begin{figure}[!htb]
  \centering
  \includegraphics[width=\columnwidth]{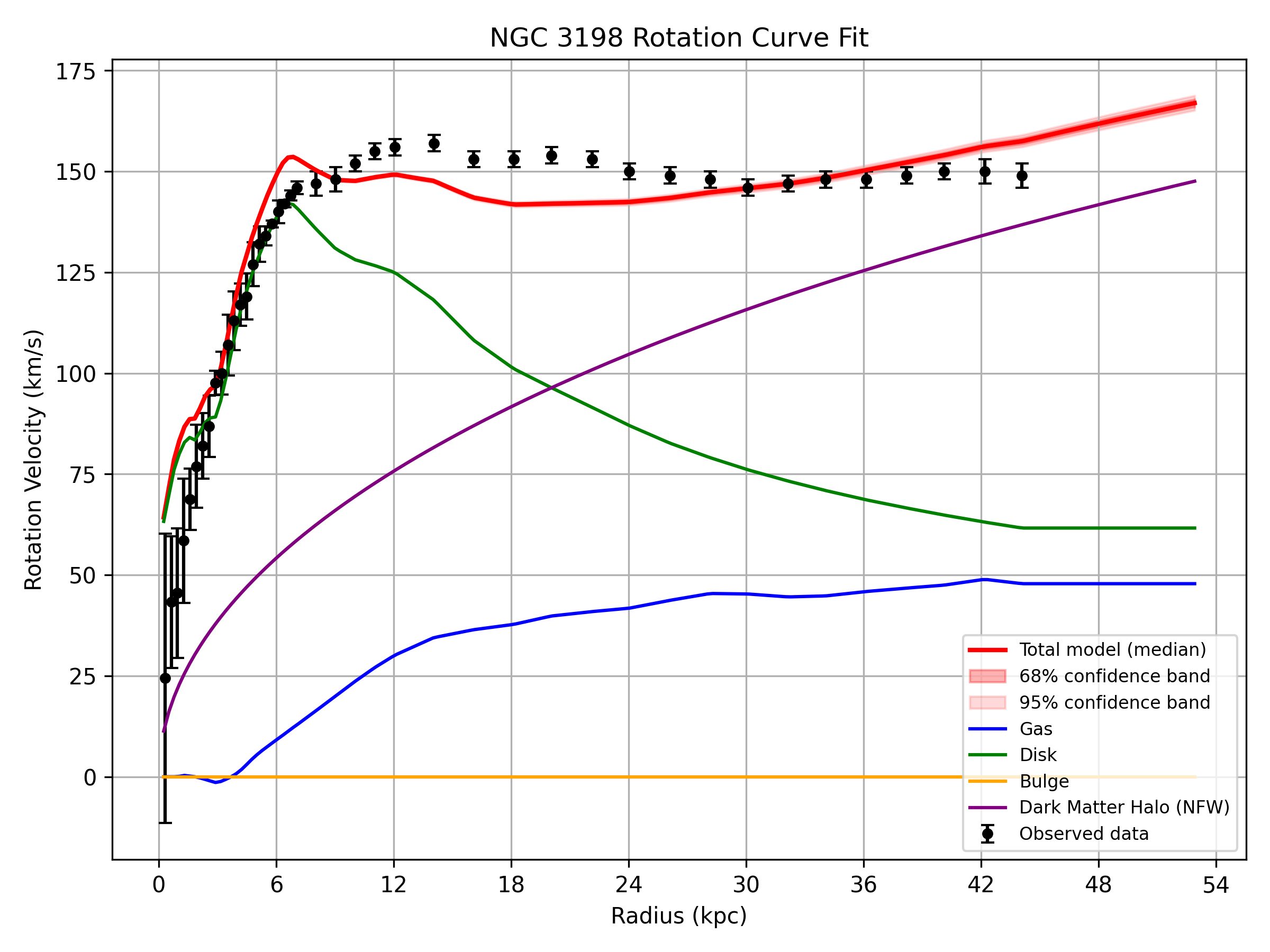}
  \caption{T2 rotation curve fit for NGC\,3198. Despite a
           visually acceptable total model (red), the
           inferred NFW halo parameters are unphysical:
           the dark matter contribution (purple) is
           entirely dominant at all radii, inconsistent
           with the observed baryonic components.}
  \label{fig:app_fit_t2}
\end{figure}

\subsection{T3 --- Exoplanet Mass--Radius Compositional}
\label{app:result_t3}

This trial (Trial~3) illustrates the inconsistent parameterisation
failure: $\log M_b = 4.939$ is reported on an inconsistent unit
scale, precluding direct comparison to the reference
$M_\mathrm{br} = 127\pm17\,M_\oplus$ \citep{muller2024mass}.
No caveat was generated by the agent.

\begin{figure}[!htb]
  \centering
  \includegraphics[width=\columnwidth]{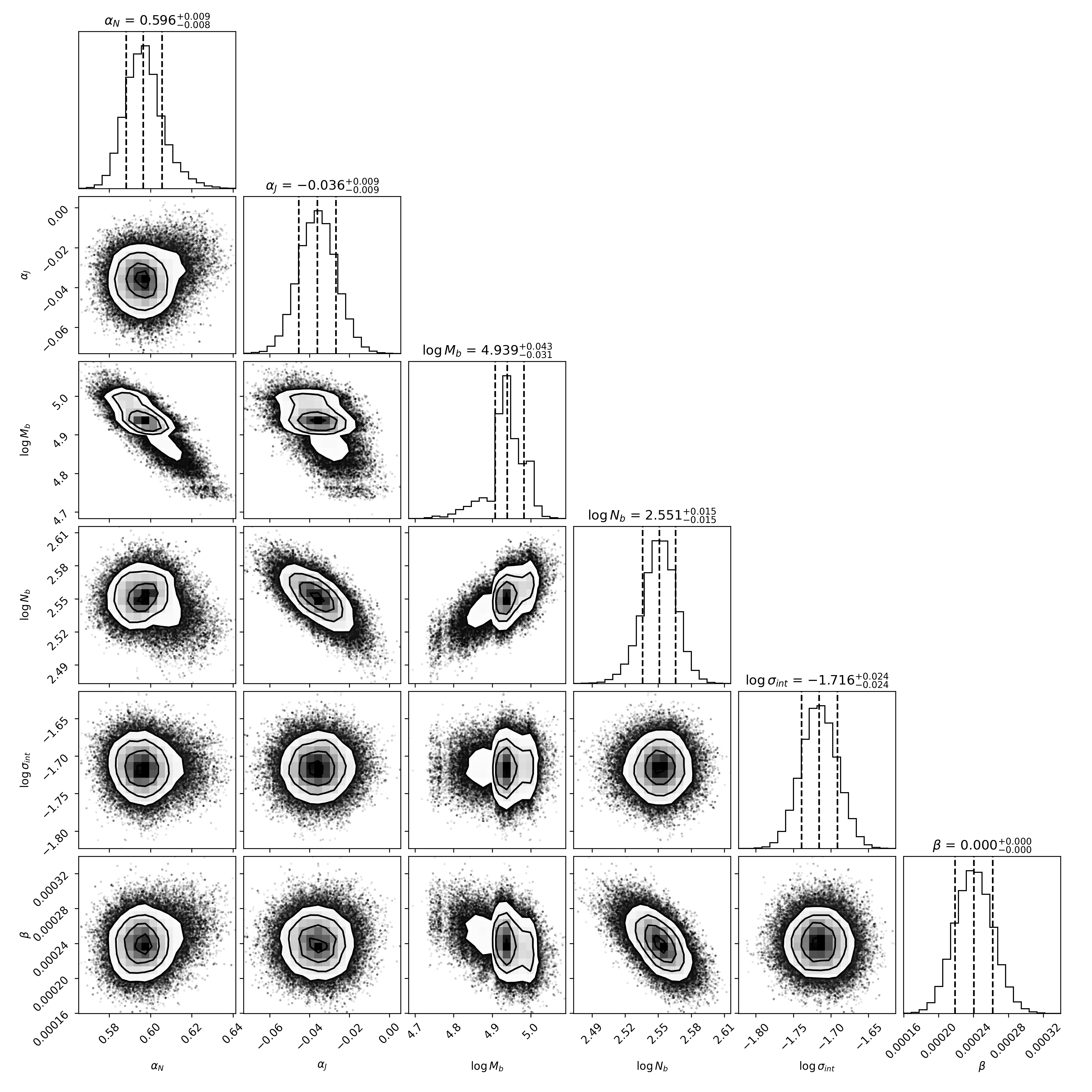}
  \caption{T3 corner plot from Trial~3. The break mass is
           reported as $\log M_b = 4.939^{+0.043}_{-0.031}$
           on an inconsistent unit scale. The inflation
           coefficient $\beta \approx 0$ is effectively
           zero, indicating the temperature-dependent
           correction was not learned. No trial flagged
           these pathologies.}
  \label{fig:app_posterior_t3}
\end{figure}

\begin{figure}[!htb]
  \centering
  \includegraphics[width=\columnwidth]{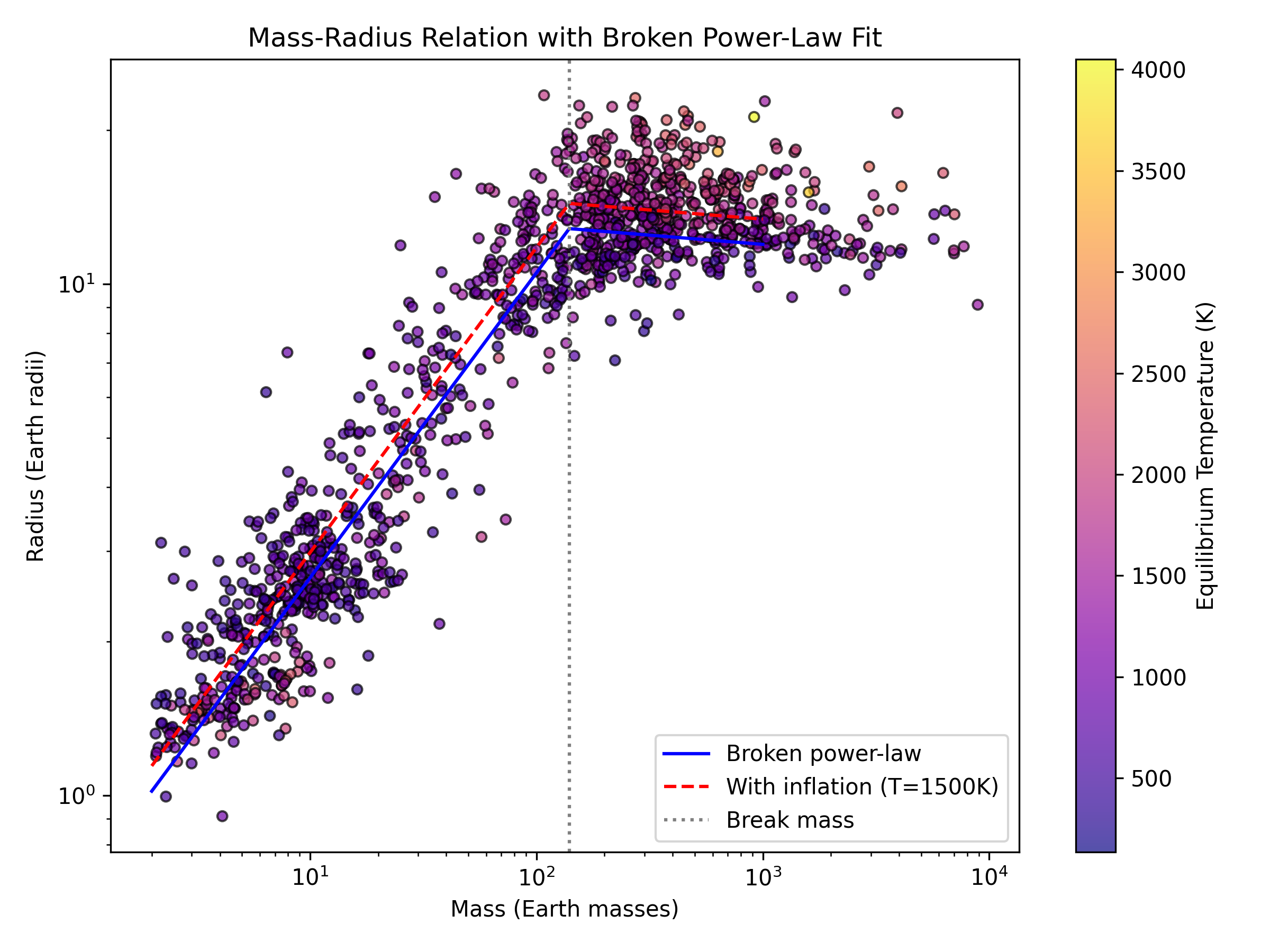}
  \caption{T3 mass--radius diagram with best-fit broken
           power-law. The fit appears visually reasonable,
           illustrating how plausible-looking outputs can
           mask underlying parameterization failures.}
  \label{fig:app_fit_t3}
\end{figure}

\subsection{T4 --- SLACS Strong Lensing (Single Completed Trial)}
\label{app:result_t4}

Of five independent trials, only one produced usable output; the
remaining four failed silently with no figures or error diagnosis.
The single completed trial shows physically coherent velocity
dispersions, but lensing-inferred values are systematically
${\approx}5\%$ above observed values ($f \approx 1.05$ vs.\
reference $1.019\pm0.008$ \citep{bolton2008sloan}), a
${\approx}4\sigma$ bias the agent did not flag.

\begin{figure}[!htb]
  \centering
  \includegraphics[width=\columnwidth]{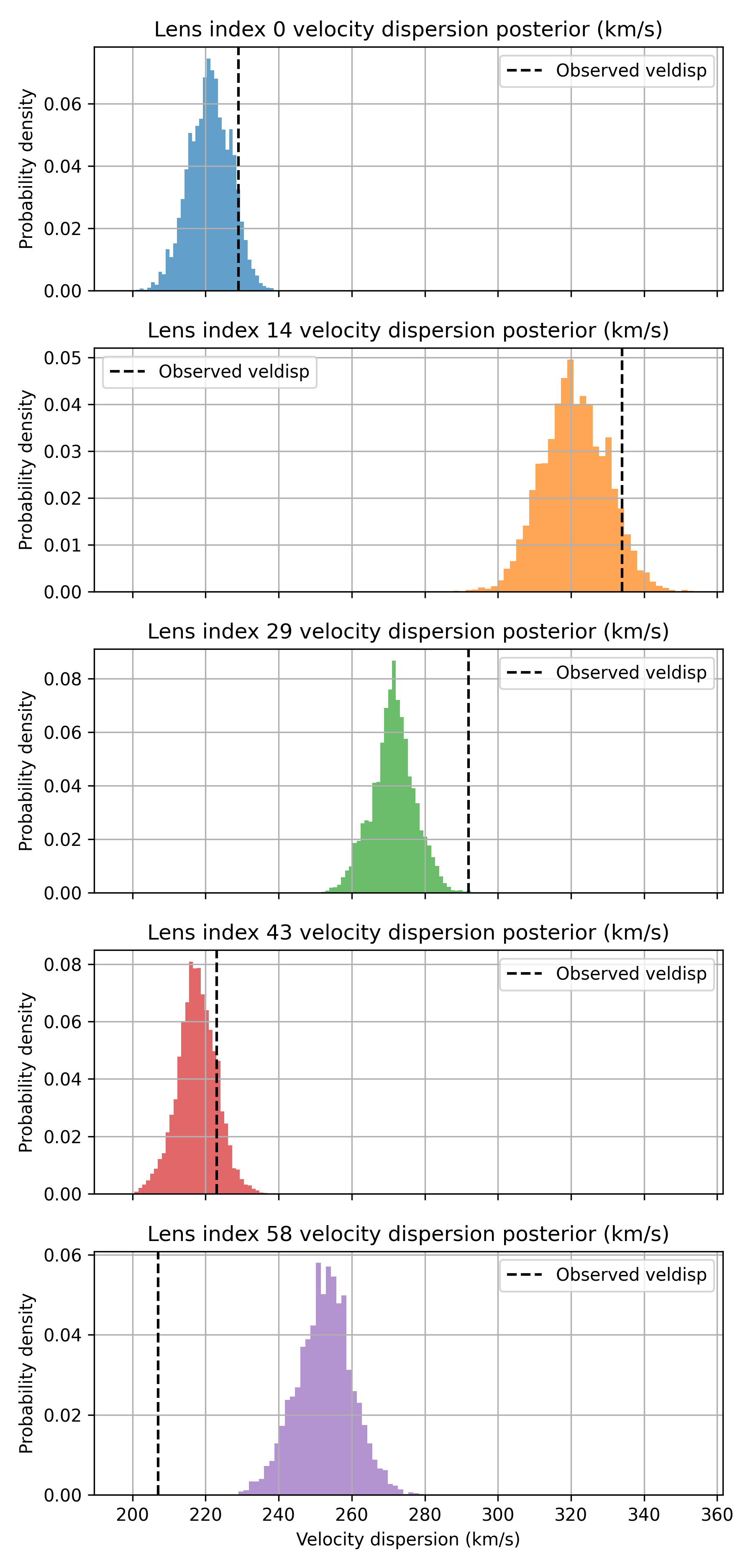}
  \caption{T4 representative 1D posteriors for five lenses.
           Several posteriors show tension with the observed
           stellar velocity dispersion (dashed line),
           consistent with the population-level ${\approx}5\%$
           systematic offset. No caveat was reported by
           the agent.}
  \label{fig:app_posterior_t4}
\end{figure}

\begin{figure}[!htb]
  \centering
  \includegraphics[width=\columnwidth]{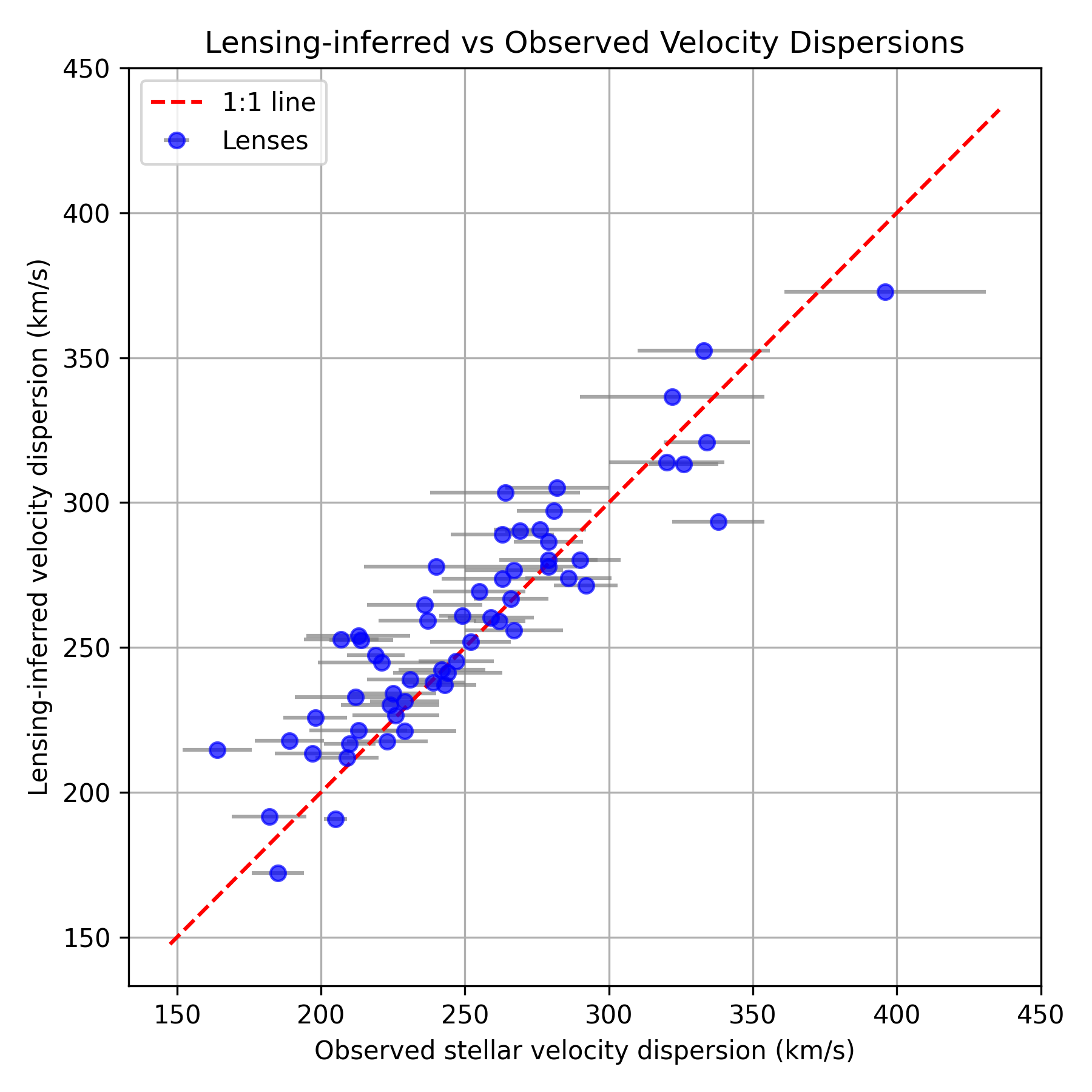}
  \caption{T4 lensing-inferred vs.\ observed stellar velocity
           dispersion $\sigma_\mathrm{SIS}$ vs.\
           $\sigma_\mathrm{obs}$ (one point per lens). Points
           lie systematically above the 1:1 line (red dashed),
           yielding $f \approx 1.05$ against reference
           $1.019\pm0.008$ \citep{bolton2008sloan}.}
  \label{fig:app_fit1_t4}
\end{figure}

\begin{figure}[!htb]
  \centering
  \includegraphics[width=\columnwidth]{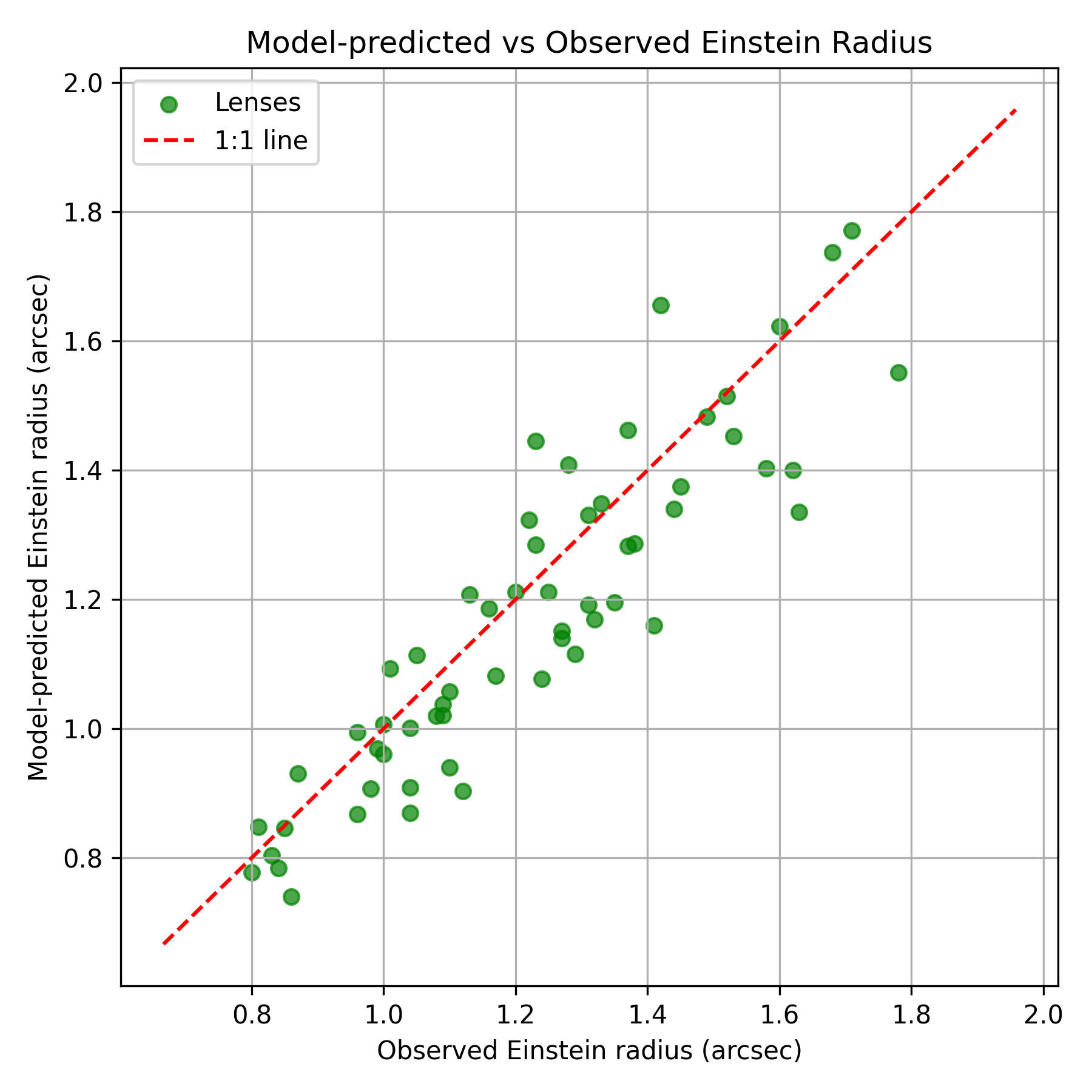}
  \caption{T4 model-predicted vs.\ observed Einstein radius
           $\theta_\mathrm{Ein}$. Agreement is closer to the
           1:1 line here than for the velocity dispersions,
           consistent with the SIS model being correctly
           implemented but the overall calibration being
           offset.}
  \label{fig:app_fit2_t4}
\end{figure}

\end{document}